\documentclass{article}
\pdfoutput=1
\usepackage{spconf,amsmath,graphicx,hyperref}
\usepackage{amssymb}
\usepackage{amsmath}
\usepackage{multirow} 
\usepackage{booktabs}
\usepackage{array}
\usepackage{xcolor}

\title{ReinPath: A Multimodal Reinforcement Learning Approach for Pathology}
 \name{Kangcheng Zhou\qquad 
       Jun Jiang \qquad 
       Qing Zhang \qquad
       Shuang Zheng \qquad 
       Qingli Li \qquad
       Shugong Xu$^{\star}$ \thanks{$^{\star}$Corresponding author} \qquad 
       }
 \address{$^{1}$East China Normal University, Shanghai, China \\
      $^{2}$Shanghai University, Shanghai, China \\
    $^{3}$Xi’an Jiaotong-Liverpool University, Jiangsu, China \\
	}
\begin{document}
\ninept
\maketitle
\begin{abstract}
Interpretability is significant in computational pathology, leading to the development of multimodal information integration from histopathological image and corresponding text data.
However, existing multimodal methods have limited interpretability due to the lack of high-quality dataset that support explicit reasoning and inference and simple reasoning process.
To address the above problems, we introduce a novel multimodal pathology large language model with strong reasoning capabilities.
To improve the generation of accurate and contextually relevant textual descriptions, we design a semantic reward strategy integrated with group relative policy optimization.
We construct a high-quality pathology visual question answering (VQA) dataset, specifically designed to support complex reasoning tasks.
Comprehensive experiments conducted on this dataset demonstrate that our method outperforms state-of-the-art methods, even when trained with only 20\% of the data.
Our method also achieves comparable performance on downstream zero-shot image classification task compared with CLIP.
\end{abstract}
\begin{keywords}
Multimodal, Reinforcement Learning, Pathology
\end{keywords}
\section{Introduction}
\label{sec:intro}
With the rapid advancement of computational pathology, foundation models have emerged as a powerful paradigm to assist pathologists in improving diagnostic accuracy \cite{li2025survey}.
In particular, multimodal pathology foundation models \cite{lu2024visual, wang2024pathology} joint analyze histopathological images and corresponding diagnostic reports, achieving cross-modal alignment between visual and textual data.
This integration enhances the model's semantic understanding of pathological features, thereby improving its performance across a wide range of downstream tasks.

Interpretability and scalability are significant in multimodal pathology models.
In a previous study, the multimodal pathology models, such as Quilt-LLaVA \cite{seyfioglu2024quilt}, Path-Asst \cite{Pathasst} and PathChat \cite{lu2024multimodal}, were trained by visual instruction fine-tuning, but this training method only allowed the model to memorize the training data, and could not form reasoning thinking chain and reasoning ability.
Apart from a good design of structure with both fine-tuning and reasoning, large scale and richly annotated dataset can also improve the model's performance and interpretability.
However, the larger scale the dataset, the larger the computational cost.
\begin{figure}[t]
\centering
  \includegraphics[width=\columnwidth]{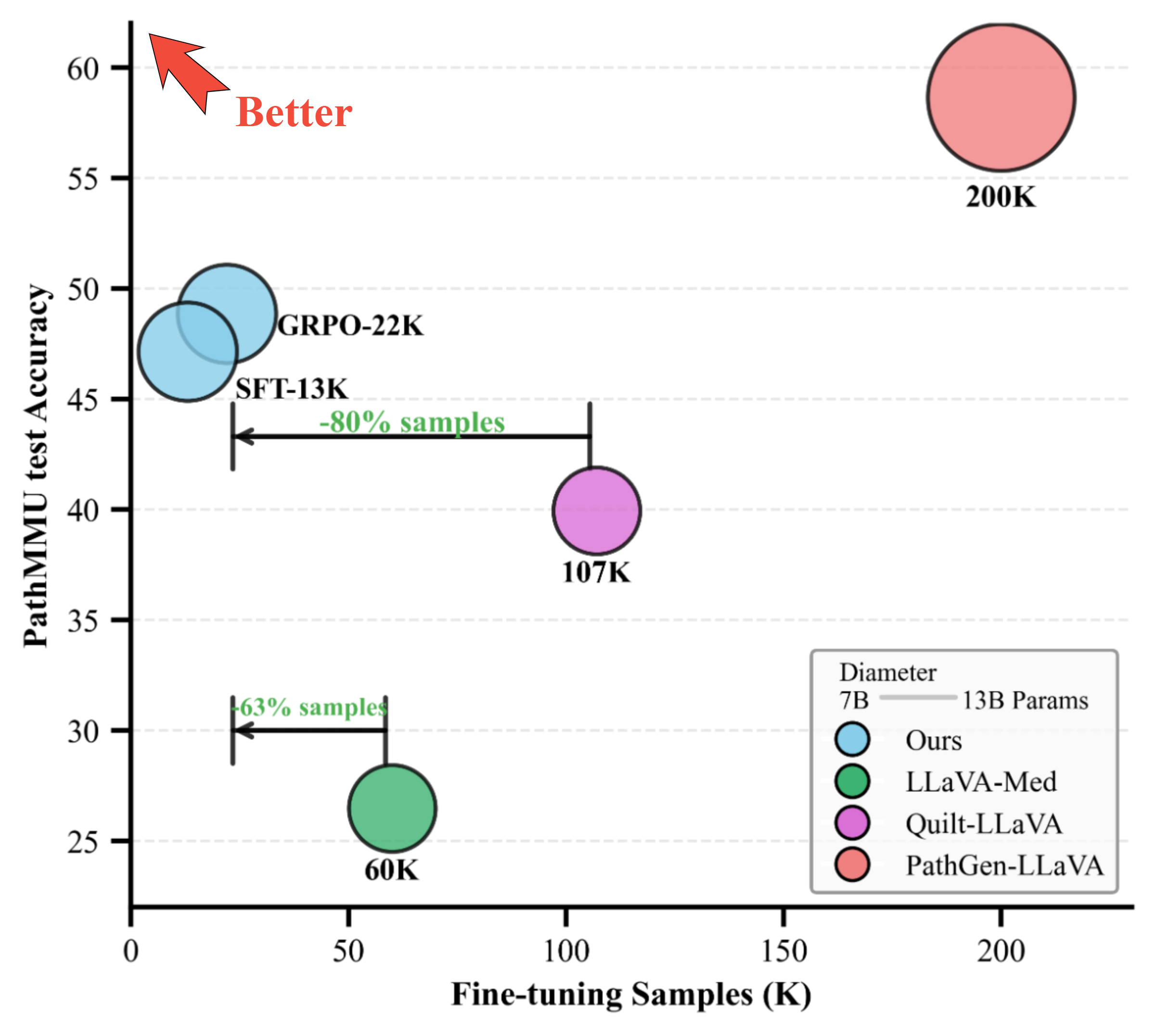}
  \caption {Performance of different fine-tuning samples and model sizes on the PathMMU dataset in VQA task.}
  \label{fig:vqa_accuracy_comparison}
\end{figure}

There have been several publicly available Pathology Visual Question Answering (VQA) datasets proposed for improving the VQA model's performance and interpretability.PathInstruct \cite{Pathasst} uses image text descriptions extracted from PubMed to generate 180K VQA data containing both detailed and conversational instructions.
Quilt-VQA \cite{seyfioglu2024quilt} parsed the data of pathology teaching videos through GPT4 to generate VQA data containing simple reasoning content.
However, these datasets lack explicit annotations of reasoning chains, i.e., explanations that connect visual findings to diagnostic conclusions.In the absence of such annotations, models struggle to develop, interpretable decision-making processes.
\begin{figure*}[t]
 \centering
  \includegraphics[width=\textwidth]{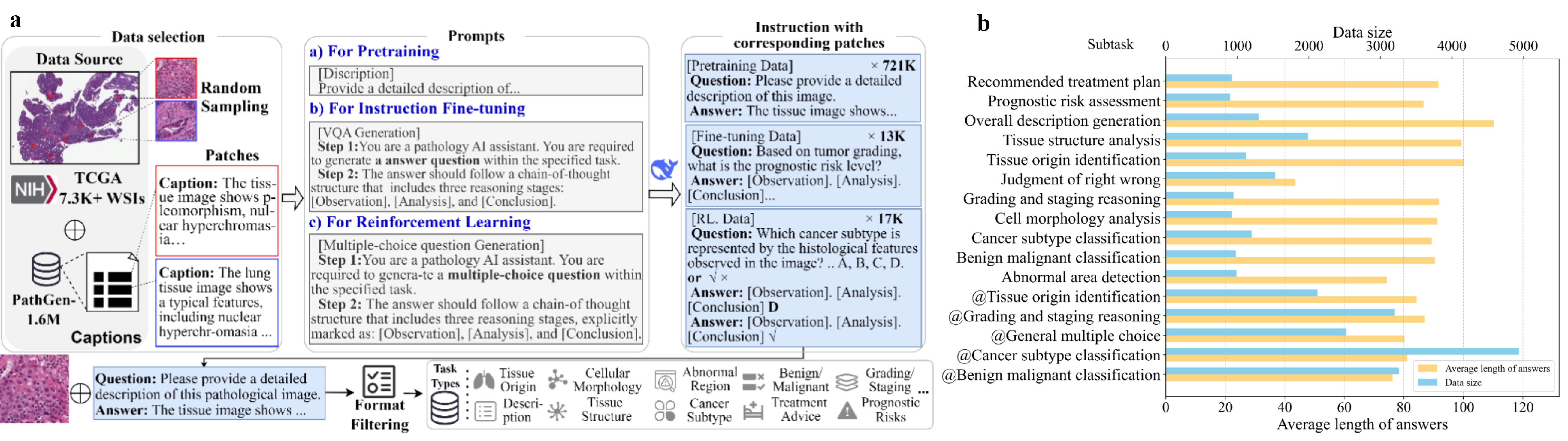}
  \caption{ReinPathVQA dataset construction pipeline (a) and the dataset's distribution (b).}
  \label{fig:dataset}
\end{figure*}
To address these challenges, we propose a novel multimodal \textbf{Rein}forcement model for \textbf{Path}ology large language model utilizing small yet informative dataset, named as \textbf{ReinPath} in Fig.~\ref{fig:vqa_accuracy_comparison}.
In this framework, we design a reinforcement learning strategy with semantic reward into the large language model to improve model's performance and interpretability.
Inspired by the Group Relative Policy Optimization (GRPO) \cite{shao2024deepseekmath}, we construct a high-quality and small-scale multimodal \textbf{Rein}forcement \textbf{Path}ology \textbf{V}isual \textbf{Q}uestion \textbf{A}nswering dataset, named as \textbf{ReinPathVQA} dataset.
As illustrated in Fig.~\ref{fig:vqa_accuracy_comparison}, our method achieves superior performance on the PathMMU dataset \cite{sun2024pathmmu}, even when trained on a smaller dataset. This result highlights the effectiveness of our approach in leveraging high-quality annotations, including detailed reasoning chains, to enhance model interpretability and performance even when trained on a smaller dataset.

Contributions of the paper are summarized as:

\begin{itemize}
   \item We construct a high-quality multimodal pathological reasoning dataset to enhance the model's interpretability, named as \textbf{ReinPathVQA}, comprising 721K pre-train samples, 13K instruction-following samples and 17K reinforcement learning samples.
   
    \item We propose a novel multimodal large language model for pathological analysis \textbf{ReinPath}, combining a multi-image encoder architecture and a reinforcement learning strategy.

    \item We introduce a semantic reward strategy to guide the generation of more accurate and contextually relevant descriptions, thereby improving both the quality and interpretability.

    \item Comprehensive experiments show that the proposed ReinPath achieves superior performance using only 20\% training data by comparable models, highlighting the effevtiveness of the ReinPathVQA dataset and ReinPath architecture, while delivering comparable performance in both open-ended VQA and zero-shot classification.
\end{itemize}
\section{Methods}
\label{sec:format}
\subsection{Dataset Construction Pipeline} 
\label{ssec:Dataset Construction Pipeline}


To construct a high-quality multimodal pathology dataset ReinPathVQA, we design a pipeline utilizing LLM based on PathGen-1.6M \cite{sun2024pathgen}, as shown in Fig~\ref{fig:dataset}.
The original 1.6M pathology image-text pairs are sourced from over 7.3k WSIs in the TCGA dataset, along with their corresponding pathology reports. 
To balance data volume and diversity, We first randomly select 100 patches and text descriptions from each WSI in PathGen-1.6M, creating a pool of candidate image-text pairs.
For each group of pairs, we then randomly select a question from a predifined set of questions to construct a basic VQA pair.

We further generate a supervised fine-tuning dataset for various sub-tasks.
To mitigate potential data overlap between the pre-training and fine-tuning stages, we randomly select 10 image-text pairs from each WSI in PathGen-1.6M.
Building upon the preprocessed dataset, we employ a LLM to independently generate question-answer pairs for each sub-task.
During prompt design, LLM plays the role of a pathology expert.
Specifically, it is guided by the provided pathological image description and visual analysis of image, to generate task-specific pathology questions and professional structured responses.
Here, the answers must follow a three-stage reasoning process, formatted as: \textit{[Observation] [Analysis] [Conclusion]}.
The generated questions are primarily short-answer type and the question categories include 4 major domains, covering a total of 10 pathology-related sub-tasks.

Additionally, we also build a reinforcement learning dataset based on task requirements.
Since reinforcement learning dataset requires strictly standardized answers, we carefully design cue prompts to guide LLM in generating multiple-choice and true/false questions that meet the requirements of specific task categories.
During the answering process, LLM is instructed to reason step-by-step based on the image description, while maintaining a consistent output format: \textit{[Observation]→ [Analysis]→ [Conclusion]}.
Fig.~\ref{fig:dataset} presents the distribution of data size and average answer length across different subtasks. 
For 10 instruction fine-tuning subtasks, the average number of instances per subtask exceeds 1000, with most responses containing more than 80 tokens.

\subsection{Multimodal Reinforcement Learning Framework}
\label{ssec:Multimodal Pathology Reinforcement Learning Framework}
\begin{figure}
 \centering
  \includegraphics[width=\linewidth]{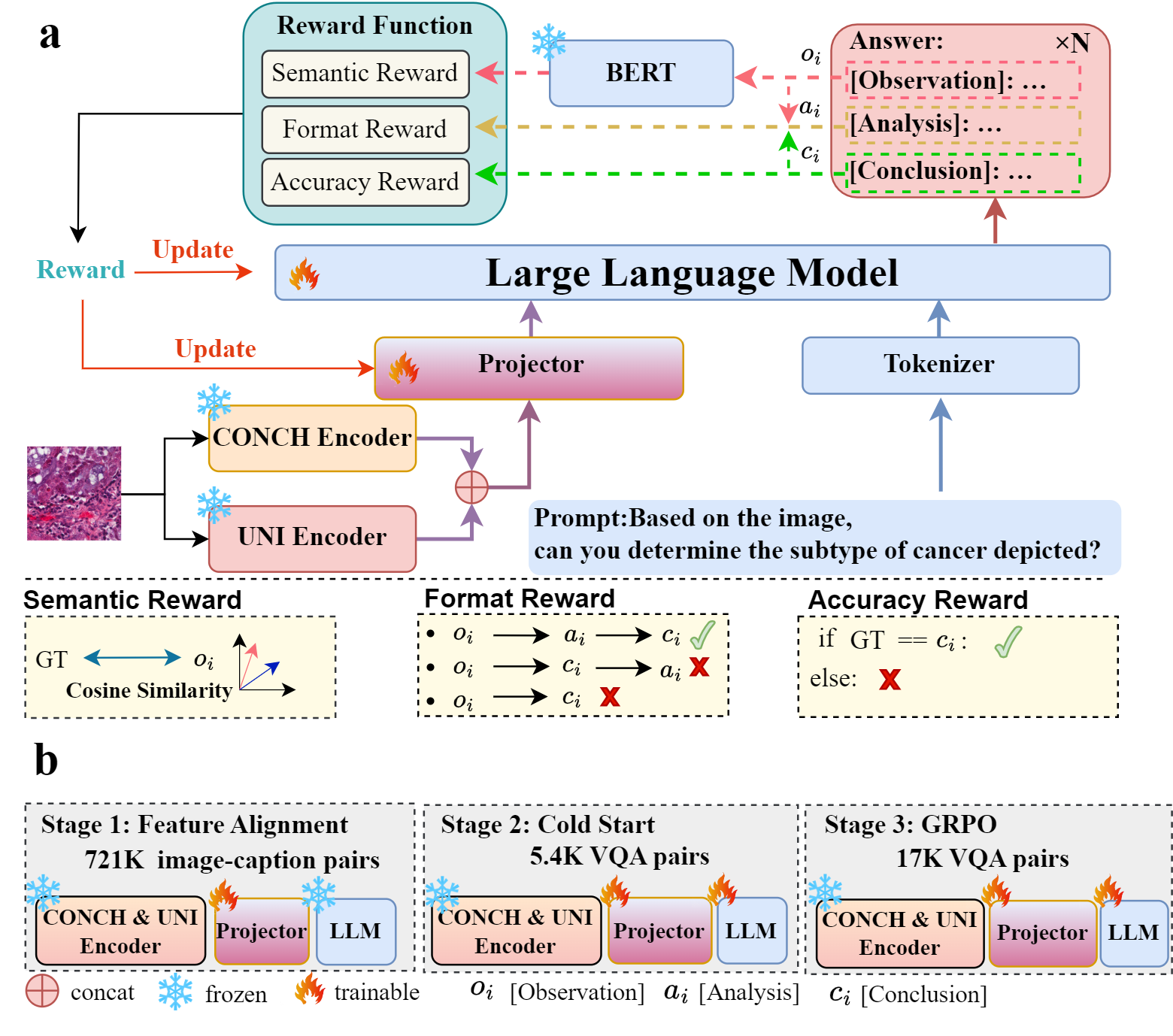}
  \caption{Overview of our ReinPath (a) with three training stages (b): Feature Alignment, Cold Start and GRPO.
  }
  \label{fig:framework}
\end{figure}

Fig.~\ref{fig:framework} shows the framework of the proposed ReinPath.
Specifically, we employ two pathological foundation model, i.e., CONCH and UNI, to extract image features.
CONCH \cite{lu2024visual} extracts more global semantic information because it has been aligned with global text information.
UNI \cite{chen2024towards} obtains fine-grained visual features as it is designed with a self-supervised manner.
The visual features are concatenated on the channel dimension as the fused visual feature, providing richer visual information for the following LLM.

To comprehensively refine the LLM-generated text, we propose an optimization strategy with three reward functions: the format reward function, the accuracy reward function, and the semantic similarity reward function.
The format reward function checks whether the model's answer follows the specified format. 
If the format conforms to \textit{[Observation]→ [Analysis]→ [Conclusion]}, the format reward $R_{for}$ is 0.5, otherwise 0.
We then introduce an accuracy reward function $R_{acc}$ to evaluate the consistency between the predicted answer $p$ and the conclusion part of the standard one $gt^*$.
\begin{equation}
  \label{eq:accuracy}
  R_{acc} = equal(gt^*, f(p)),
\end{equation}
where $equal$ is to judge if the the variations are consistent,the consistent value is 2, otherwise 0,and $f$ represents a regular express to extract conclusion of $p$.
Apart from the format and accuracy requirements, merging semantic information of image into the text description is also significant, especially during model's reasoning process. Therefore, we propose a semantic similarity reward function to assess the congruence between the generated descriptive content and the standard answer.We utilize a frozen text encoder pre-trained on medical texts to obtain text features of the standard answer description and the observed part of the predicted one. Then, we calculate their cosine similarity as the semantic reward.
The formula is as follows:
\begin{equation}
  \label{eq:semantic}
    R_{sem} = S(E(gt), E(o)),
\end{equation}
where $S$ is the computed cosine distance, $E$ represents the text encoder, $o$ is the observed part of the model output,and $gt$ stands for the observation part in the standard answer.
Given a question $q$, GRPO generates $N$ candidate responses $\{d_1,d_2,... ,d_N\}$, and use the reward function $R$ to evaluate each answer $d_i$. Specifically, the three parts of the reward are calculated separately for each answer $d_i$ and then added together to obtain the initial reward for each answer $d_i$.The formula is as follows:
\begin{equation}
  \label{eq:all_reward}
    R = R_{for}+R_{acc}+R_{sem},
\end{equation}
GRPO normalizes the reward $R$ by calculating the mean and standard deviation, and then obtains the final reward:
\begin{equation}
  \label{eq:relative advantage}
A_i = \frac{r_i - mean\{r_1, r_2, \ldots, r_N\}}{std\{r_1, r_2, \ldots, r_N\}},
\end{equation}
where $A_i$ represents the advantage of candidate answer $d_i$ over other sampled responses, and $r_i$ represents the original reward $R$ for answering $d_i$. GRPO encourages the model to generate responses with higher rewards within the group by updating the policy $\pi_{\theta}$ using formula:
\begin{equation}
\label{eq:GRPO2}
J_{GRPO}(\theta) = \mathbb{E}[\{d_i\}_{i = 1}^N \sim \pi_{\theta_{old}}(q)],
\end{equation}
\begin{equation}
\label{eq:GRPO3}
\frac{1}{N}\sum_{i = 1}^{N} \left\{ \min[s_1 \cdot A_i, s_2 \cdot A_i] - \beta \mathbb{D}_{KL}[\pi_{\theta}||\pi_{ref}] \right\},
\end{equation}
\begin{equation}
\label{eq:GRPO4}
s_1 = \frac{\pi_{\theta}(d_i|q)}{\pi_{\theta_{old}}(d_i|q)},
\end{equation}
\begin{equation}
\label{eq:GRPO5}
s_2 = \text{clip}\left(\frac{\pi_{\theta}(d_i|q)}{\pi_{\theta_{old}}(d_i|q)}, 1 + \epsilon, 1 - \epsilon \right).
\end{equation}


\subsection{Training Strategy}
\label{ssec:Training Strategy}
To comprehensively optimize the our model, we design a three-stage training strategy. Stage 1 focuses on image-text feature alignment by projecting visual features into the semantic space of text features through a two-layer MLP on the constructed 721k ReinPathVQA dataset. During training, only the projector is optimized, while both the LLM and the image encoder are kept frozen.

Stage 2 is designed to accelerate convergence and standardize the output format by leveraging a cold start training. 
It uses 5.4k high-quality instruction tuning samples selected from 13K SFT data for long thought chains and uniform category distribution to capture complex visual-linguistic patterns.
Training follows a standard visual instruction tuning paradigm, with the projector and LLM trained and the image encoder frozen.

Following the cold start training, we acquire basic instruction following capabilities but still lacks sufficient reasoning ability.
Therefore, Stage 3 solves this problem by continuing training with 17k reinforcement learning samples, enhancing the model's capacity for multi-step reasoning and beter interpretability.
For each input question, we generate 4 candidate answers, each evaluated with the above 3 reward functions.
These comprehensive reward functions are normalized by Eq.~\ref{eq:relative advantage} and then utilized for updating the policy model.During this stage, only the LLM and the projector are trainable, while the image encoder remains frozen.
\begin{table}[t]
    \centering

    \resizebox{\columnwidth}{!}{
    \begin{tabular} {
  >{\centering\arraybackslash}m{2.3cm}       
  >{\centering\arraybackslash}m{0.7cm}
  >{\centering\arraybackslash}m{0.9cm}|
  >{\centering\arraybackslash}m{0.45cm}
  >{\centering\arraybackslash}m{0.45cm}|
  >{\centering\arraybackslash}m{0.45cm}
  >{\centering\arraybackslash}m{0.45cm}|
  >{\centering\arraybackslash}m{0.8cm}
  }
    \hline
    \multirow{2}{*}{Models} & 
    \multirow{2}{*}{Params} & 
    \multirow{2}{*}{Samples} & 
    \multicolumn{2}{c|}{Quilt-VQA} & 
    \multicolumn{2}{c|}{PathVQA} & 
    PMC-VQA \\
     &  &  & Recall & Acc & Recall & Acc & Acc  \\ 
     \cmidrule(lr){1-8}
    LLaVA-1.5 \cite{liu2024improved} & 13B & 665K & 58.8 & \underline{67.7} & 11.7 & 54.0 & 35.1 \\
    LLaVA-Med \cite{li2023llava} & 7B & 60K & 54.8 & 61.2 & 12.0 & 56.2 & 1.3 \\
    Quilt-LLaVA \cite{seyfioglu2024quilt} & 7B & 107K & \textbf{60.9} & 64.4 & \textbf{15.3} & \textbf{58.7} & 32.4 \\
    SFTPath & 8B & 13K & 48.4 & 58.3 & 13.3 & \underline{57.0} & \underline{38.1} \\
    ReinPath & 8B & 22K & \underline{60.1} & \textbf{72.9} & \underline{14.3} & 54.4 & \textbf{38.6} \\ 
    \hline
  \end{tabular}
  }
  \caption{Comparison of performance on three VQA datasets in terms of recall and accuracy (\%).
  \textbf{Bold} means the best results and \underline{underline} is the second-best.
  }
  \label{tab:VQA1}
\end{table}

\section{Experiments}
\label{sec:Experiments}
\subsection{Implementation Details}
In this paper, we adopt Llama3-8B as LLM. We incorporate a two-layer MLP as a projector for modal alignment. We train two models with SFT (only 2 stages) and GRPO by LoRA, named SFTPath and ReinPath. The rank of LoRA is set to 128. The learning rate is set to 2e-4 with a cosine learning rate scheduler and a batch size of 32. Deepspeed Zero3 is used as the training framework.We conduct experiments on a server equipped with 4 Nvidia A6000 GPUs.
\begin{table}[t]
  \centering
  \small
  \fontsize{9}{10}\selectfont
  \begin{tabular}{
  >{\centering\arraybackslash}m{2.4cm}       
  >{\centering\arraybackslash}m{0.8cm}
  >{\centering\arraybackslash}m{0.5cm}
  >{\centering\arraybackslash}m{0.5cm}
  >{\centering\arraybackslash}m{0.8cm}
  >{\centering\arraybackslash}m{0.8cm}}
\hline
Models & PubMed & EduC & Atlas & PathCLS &Overall    \\ 
 \cmidrule(lr){1-6}
\multicolumn{6}{c}{General MLLMs}                \\ 
 \cmidrule(lr){1-6}
LLaVA-1.5-13B   & 41.0   & 39.4      & 44.3  & 23.5    & 37.6         \\
Qwen-VL-MAX & \underline{50.9} & 47.9 & 49.8 & 29.6 & 45.9 \\
Gemini ProVision  & 44.9   & 43.7   & 49.4  & 34.7  & 42.7    \\ 
 \cmidrule(lr){1-6}
\multicolumn{6}{c}{Pathology-Specific MLLMs}  \\ 
\cmidrule(lr){1-6}
LLaVA-Med  & 27.7 & 27.2 & 30.7 & 20.3 & 26.2 \\
Quilt-LLaVA  & 42.6   & 45.3    & 42.7  & 29.2    & 41.5        \\
PathGen-LLaVA & \textbf{60.1} & \textbf{60.7} & \textbf{64.9} & \textbf{48.9} & \textbf{58.4} \\
SFTPath  & 47.8   & 49.4       & 55.8  & 35.6    & 47.2          \\
ReinPath & 49.9 & \underline{49.6} & \underline{57.4} & \underline{38.5} &  \underline{48.9} \\ 
\hline
  \end{tabular}
  \caption{Performance comparison on the PathMMU dataset.}
  \label{tab:VQA2}
\end{table}

\begin{table}[t]
  \centering
  \small
  \fontsize{9}{10}\selectfont
  \begin{tabular}{>{\centering\arraybackslash}m{2.5cm}
                  >{\centering\arraybackslash}m{0.6cm}
                  >{\centering\arraybackslash}m{0.6cm}
                  >{\centering\arraybackslash}m{0.6cm}
                  >{\centering\arraybackslash}m{1.8cm}
                  }
    \hline
    Models & Lung & Colon & CRC & WSSS4LUAD  \\
    \cmidrule(lr){1-5}
    \multicolumn{5}{c}{CLIP Models} \\
    \cmidrule(lr){1-5}
    OpenAI CLIP \cite{CLIP} & 31.5 & 75.7 & 22.2 & 61.6  \\
    PLIP \cite{PLIP} & 86.0 & 87.0 & 53.1 & 69.5  \\
    PathCLIP \cite{Pathasst} & 88.7 & 94.3 & 54.2 & 81.1 \\
    PathGen-CLIP \cite{sun2024pathgen} & 90.0 & 97.5 & 63.3 & 82.9 \\
     \cmidrule(lr){1-5}
    \multicolumn{5}{c}{Pathology-Specific Multimodal Large Language Models} \\
    \cmidrule(lr){1-5}
    LLaVA-Med \cite{li2023llava} & - & - & 22.2 & - \\
    Quilt-LLaVA \cite{seyfioglu2024quilt} & 17.5 & 44.0 & 11.1 & 78.2 \\
    ReinPath & 62.1 & 95.2 & 38.4 & 83.2 \\
    \hline
    \end{tabular}
  
  \caption{Zero-shot image classification results.}
  \label{tab:classfy}
\end{table}

\subsection{Experiment Results}
\label{ssec:Experiment Results}
We conduct experiments on the 4 public VQA datasets:Quilt-VQA\cite{seyfioglu2024quilt}, Path-VQA\cite{he2020pathvqa}, PMC-VQA\cite{zhang2023pmc} and PathMMU\cite{sun2024pathmmu}. As shown in Table~\ref{tab:VQA1}, our proposed model achieves outperform performance across multiple pathology VQA benchmarks. Notably, ReinPath achieves the highest accuracy on both Quilt-VQA and PMC-VQA datasets, surpassing existing baseline methods by 5.2\% and 3.5\%, respectively. Notably, ReinPath only utilizes a 22K high-quality training dataset.

As shown in Table~\ref{tab:VQA2}, ReinPath achieves an overall score of 48.9\%, closely following the SOTA model. Compared to PathGen-LLaVA, our model uses only one-tenth of the fine-tuned data, yet manages to achieve about 80\% performance, highlighting the effectiveness of our approach.

We select three publicly available histopathological image benchmarks to evaluate the performance on the zero-shot image classification task:LC-Lung and LC-Colon\cite{borkowski2019lung}, CRC\cite{kather2018100} and WSSS4LUAD\cite{han2022wsss4luad}.As shown in Table~\ref{tab:classfy}, our model achieves the best performance among all pathology-specific MLLMs in zero-shot image classification. On the WSSS4LUAD dataset, we even surpass the best-performing CLIP variant by 0.3\%. Notably, while CLIP-based models are trained specifically for image-text contrastive learning and excel in classification tasks, our model is designed to support both classification and open-ended VQA. This demonstrates that our model can achieve strong performance on classification tasks while maintaining the flexibility required for complex downstream applications.

\begin{figure}[t]
 \centering
  \includegraphics[width=\columnwidth]{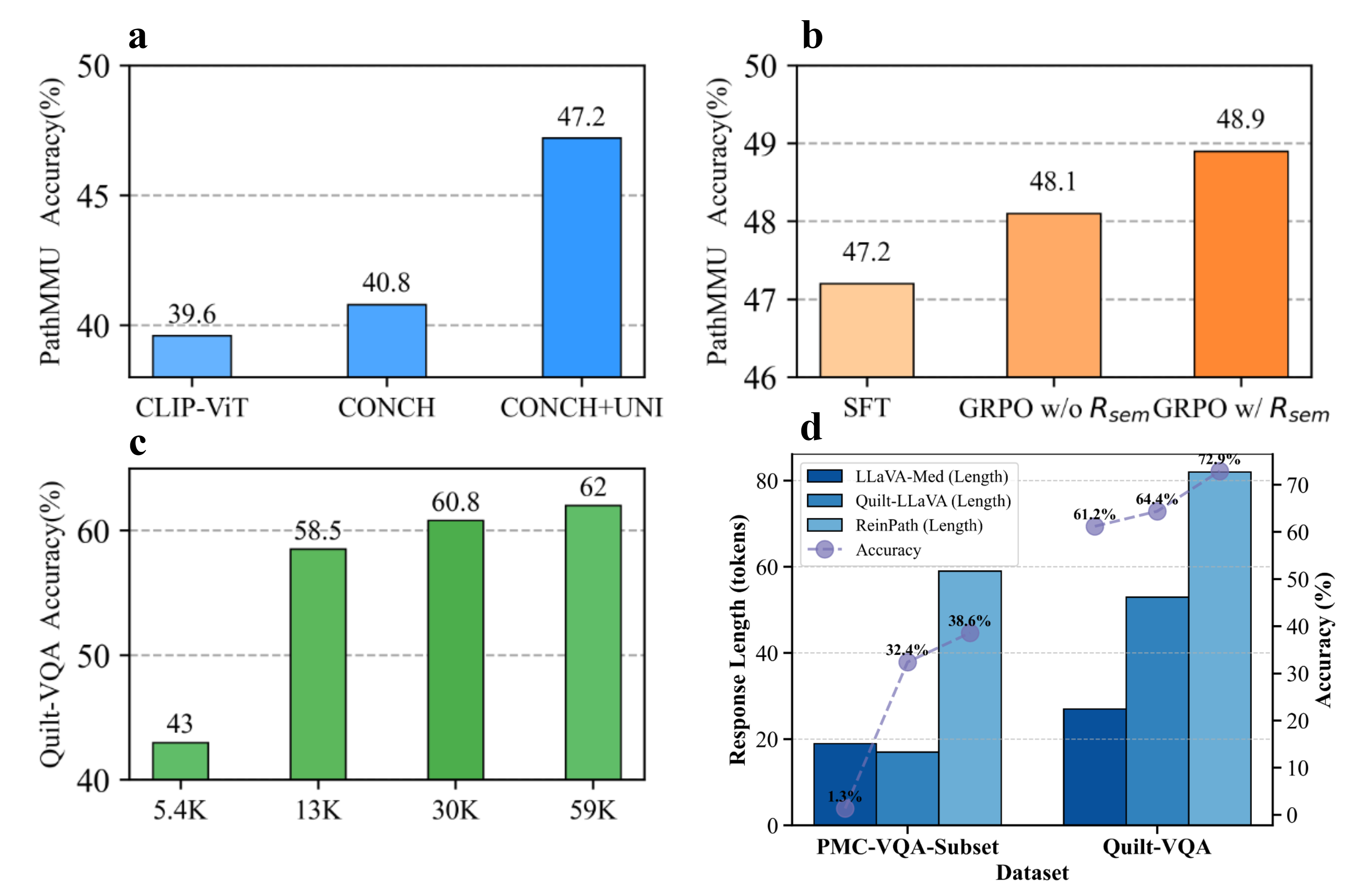}
  \caption{Ablation experiments with encoder (a), training strategy (b) and SFT data size (c). Comparison of model response length and accuracy on two VQA datasets (d).}
  \label{fig:ablation}
\end{figure}

To evaluate the enhancement of model interpretability, we compared  ReinPath  with  baseline  across  VQA datasets, focusing on their answer lengths and accuracy. As illustrated in Fig.~\ref{fig:ablation}, ReinPath generates longer answers with step-by-step explanations, enhancing the clarity and interpretability of its reasoning, thus improving performance. Compared with the baseline model, our model reasoning through normal reasoning logic, and can give the correct answer through reasoning.
\begin{table}[t]
  \centering
  \small
\begin{tabular}{
  >{\centering\arraybackslash}m{1.4cm}
  >{\centering\arraybackslash}m{0.5cm}
  >{\centering\arraybackslash}m{1.8cm}
  >{\centering\arraybackslash}m{0.4cm}
  >{\centering\arraybackslash}m{2.1cm}}
\hline
\multicolumn{2}{c}{Pretrain dataset} & \multicolumn{2}{c}{SFT dataset} & \multirow{2}{*}{
\begin{tabular}[c]{@{}c@{}}Quilt-VQA\\ ACC\end{tabular}} \\ 
\cmidrule(lr){1-2} \cmidrule(lr){3-4}
Quilt-1M & 721K & Quilt-107K & 13K &  \\ 
 \cmidrule(lr){1-5}
\checkmark &  & \checkmark &  & 44.3\% \\
\checkmark &  &  & \checkmark & 56.4\%(\textbf{+12.1\%}) \\
 & \checkmark &  & \checkmark & 60.2\%(\textbf{+15.9\%}) \\ \hline
\end{tabular}%
 
  \caption{Ablation experiments comparing the effectiveness of different training datasets with CONCH as image encoder.
  }
  \label{tab:Ablation1}
\end{table}



\subsection{Ablation Study}
\label{ssec:Ablation Study}

As shown in Table~\ref{tab:Ablation1}, replacing the SFT dataset with our generated dataset leads to a significant improvement of 12.1\%. This substantial gain highlights its superior annotation quality and rich semantic content. Furthermore, when we pretrain the model on our large-scale 721K dataset, the performance further improves by 3.8\%. This demonstrates that our dataset is of high quality and contributes meaningfully to learning robust, generalizable representations.
Additionally, scaling experiments in Fig.~\ref{fig:ablation}. As the dataset size increases, so does the accuracy. These emphasize the high quality and scalability benefits of ReinPathVQA in enhancing model performance.

As shown in Fig~\ref{fig:ablation}, our dual image encoder architecture significantly enhances performance in visual question answering. Meanwile, using GRPO improves the accuracy to 48.1\%, while further integrating semantic rewards $R_{sem}$ into GRPO leads to a notable enhancement, reaching an accuracy of 48.9\%. This demonstrates that our proposed method, which combines GRPO with semantic rewards $R_{sem}$, effectively enhances model performance.



\section{Conclusion}
\label{sec:Conclusion}
In this work, we introduce a high-quality multimodal pathology dataset, \textbf{ReinPathVQA}, to enhance model interpretability through detailed reasoning annotations. Based on this dataset, we propose a novel multimodal pathology large language model, \textbf{ReinPath}, introducing a semantic reward into the GRPO to improve the accuracy and relevance of generated descriptions. Our experiments demonstrate that the model achieves strong performance on both VQA and zero-shot classification tasks, even when trained on a limited amount of data. Most importantly, our approach significantly enhances the interpretability of decision-making in computational pathology, bringing greater transparency and reliability to clinical applications.



\bibliographystyle{IEEEbib}
\bibliography{strings,Template}

\end{document}